\RequirePackage[svgnames]{xcolor}

\documentclass[11pt,letterpaper]{mystyle}
\usepackage[comma,authoryear,compress]{natbib}
\usepackage{wrapfig}
\usepackage[T1]{fontenc}
\usepackage{url}
\usepackage{fontawesome5}     

\newcommand{\abstractfooter}{%
  \par\vspace{3mm}\noindent
  \parbox{\linewidth}{%
    \raisebox{-0.05em}{\faEnvelope}\hspace{0.3em}\textbf{Correspondence:} \texttt{\{xiao\_pu,ericxwang\}@ucsb.edu}\\
    \raisebox{-0.1em}{\faGithub}\hspace{0.3em}\href{https://github.com/SophiaPx/survive-or-collapse}{\texttt{github.com/SophiaPx/survive-or-collapse}}%
  }%
}

\title{\textbf{Survive or Collapse}: The Asymmetric Roles of\\ Data Gating and Reward Grounding in Self-Play RL}
\runningtitle{Survive or Collapse: The Asymmetric Roles of Data Gating and Reward Grounding in Self-Play RL}

\newcommand{\ucsb}{1}
\newcommand{\cisco}{2}

\author[\ucsb]{Sophia Xiao Pu}
\author[\ucsb]{Zhaotian Weng}
\author[\ucsb]{Chengzhi Liu}
\author[\cisco]{Jayanth Srinivasa}
\author[\cisco]{Gaowen Liu}
\author[\ucsb]{\\ William Yang Wang}
\author[\ucsb]{Xin Eric Wang}

\affil[\ucsb]{University of California, Santa Barbara}
\affil[\cisco]{Cisco Research}

\begin{document}

\begin{abstract}
Self-play reinforcement learning trains language models on their own generated tasks, co-evolving a proposer and solver without human labels. Recent systems report strong reasoning gains, but collapse and instability are widely observed and poorly understood. The dominant response treats this as a reward-design problem. We argue instead that self-play stability is governed by two distinct levers: a data-level \emph{gate} that decides which proposer-generated tasks enter the training pool, and the \emph{reward signal} that updates the policy on tasks already admitted. Through controlled experiments on a Python output-prediction task and a deterministic-DSL twin task that strips pretraining priors, output ambiguity, and executor noise, we find the two levers are asymmetric. A strict gate is sufficient for stability under every reward variant we test, including a self-consistency reward with no access to ground truth; while no reward variant is sufficient once the gate is removed. This asymmetry exposes a counter-intuitive coupling we call the \emph{Grounded Proposer Paradox}: a proposer with ground-truth access accelerates collapse \emph{faster} than an ungrounded one when paired with a self-consistency solver, by concentrating training on clean tasks that form the fastest path to a spurious self-consistent attractor. Replacing the binary gate with a continuous strictness parameter $\varepsilon$ further reveals a two-stage phase transition: training-side metrics decouple at low $\varepsilon$, while validation accuracy holds until $\varepsilon$ is much higher. Data-level gating, not reward calibration, is the binding constraint on self-play stability.
\abstractfooter
\end{abstract}

\maketitle

\begin{figure}[h]
\centering
\includegraphics[width=\linewidth]{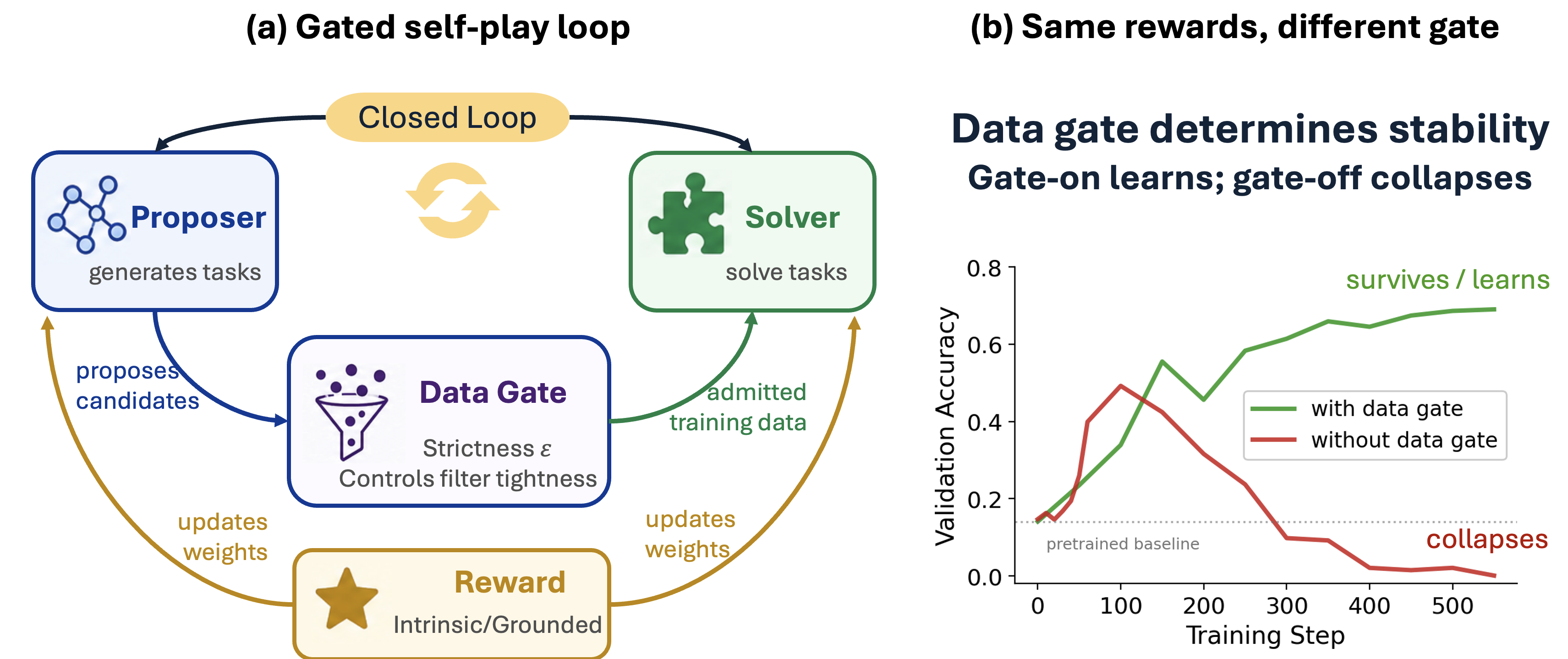}
\caption{\textbf{Same rewards, different gate, opposite outcomes.}
\textbf{(a)} In closed-loop self-play, the proposer generates candidate tasks for the solver, both updated via RL. A data gate decides which tasks enter the training pool.
\textbf{(b)} Under the same intrinsic proposer and solver rewards, gate-on training improves over the baseline, whereas gate-off training collapses below it.
}
\label{fig:teaser}
\end{figure}

\section{Introduction}

Self-play reinforcement learning with verifiable rewards is reshaping how reasoning capability is acquired in large language models. In a typical self-play loop, a proposer generates candidate problems, a solver attempts to answer them, the answers are checked against an executable verifier, and both roles are updated via reinforcement learning on the resulting reward (see Figure~\ref{fig:teaser}(a) for a schematic overview). The loop closes the supervision pipeline: the system manufactures its own training curriculum and grades its own attempts, without human-labeled tasks. The Absolute Zero paradigm demonstrates this principle \citep{zhao2025absolute}, and recent extensions adapt the loop to retrieval \citep{lu2025search}, long-context reasoning \citep{yang2025spell}, and variational task synthesis \citep{liang2025beyond}.

The reported gains are striking, but so are the failure modes. Training collapse is widely observed across self-play and self-training systems: proposers learn to emit degenerate tasks that exploit the training objective \citep{bailey2026scaling}, intrinsic rewards fail catastrophically when the model's prior confidence does not align with correctness \citep{he2026howfar}, and self-rewarding setups exhibit pervasive instability \citep{shafayat2025srt, liu2026selfplay}. The dominant response treats stability as a \emph{reward design} problem: momentum anchors, confidence penalties, and other corrections to the reward signal \citep{xu2026ace, wu2026rebound}, alongside analyses of how pretraining priors interact with intrinsic rewards \citep{he2026howfar}. The implicit assumption is that the right reward, applied to the data the proposer happens to generate, is what self-play stability requires.

We argue that this assumption misidentifies the \emph{binding constraint}. A self-play system has two distinct levers that prior work tends to conflate. The first is a \emph{data-level gate} that decides which proposer-generated tasks are admitted to the training pool. The second is the \emph{reward signal} that shapes how the policy updates on tasks already admitted. The gate decides what data exists; the reward decides how the optimizer reacts to it. Both, in principle, could govern stability.

Through controlled experiments on a Python output-prediction task and a synthetic twin task built on a deterministic 15-operator DSL that removes pretraining priors, output ambiguity, and executor noise by construction, we show that the two levers are \emph{asymmetric} (Figure~\ref{fig:teaser}). With a strict gate, every reward variant we tested remains stable, including a self-consistency reward with no access to ground truth. Without the gate, no reward variant we tested is sufficient: even a reward grounded in execution truth collapses on tasks with ambiguous outputs. \textbf{The gate is \emph{sufficient} for stability on its own; no reward variant is sufficient without it.} This is not a trivial consequence of training on low-quality data. One might expect the reward signal to naturally down-weight corrupted samples: bad data should receive low reward and contribute little to the gradient. Our experiments show this expectation is wrong. Under a self-consistency reward, the intrinsic-grounded gap (Section~\ref{subsec:gate-as-constraint}) reveals that corrupted data receives \emph{higher} reward than clean data once the solver converges to a spurious consensus, because intra-group agreement is easiest to maximize on ambiguous tasks. The reward does not down-weight bad data; it is maximized by it. Calibrating the reward while leaving the data pipeline open (the dominant approach in the current literature) addresses a downstream symptom rather than the binding constraint.

Three findings follow. \emph{First}, removing the gate causes irreversible collapse regardless of reward design. With a self-consistency solver reward, the policy converges to a self-consistent attractor that is decoupled from correctness; we quantify this through the \emph{intrinsic-grounded gap}, the difference between the solver's self-consistency reward and its grounded accuracy. The collapse replicates on the DSL twin, establishing it as a property of the optimization dynamics rather than the coding environment. \emph{Second}, we identify a counter-intuitive coupling between the two roles, the \emph{Grounded Proposer Paradox}: a proposer with access to ground-truth verification accelerates collapse \emph{faster} than an ungrounded one when paired with a self-consistency solver, because it concentrates output on clean, well-structured tasks that form the lowest-resistance path to the spurious attractor. \emph{Third}, replacing the binary gate with a continuous strictness parameter $\varepsilon \in [0, 1]$ reveals a sharp phase transition with a two-stage structure: training-side metrics decouple at low $\varepsilon$, but downstream validation accuracy remains stable until $\varepsilon$ is substantially higher.

Section~\ref{sec:formalization} formalizes the self-play setup, the two axes, and the DSL control environment. Section~\ref{sec:policy-collapse} dissects the collapse mechanism, the Grounded Proposer Paradox, and the gate's role as the binding constraint. Section~\ref{sec:phase-transitions} maps the phase transition under continuous gate relaxation. Section~\ref{sec:related-work} surveys related work and Section~\ref{sec:conclusion} concludes.

\section{Formalizing the Asymmetric Dynamics in Self-Play}
\label{sec:formalization}
\subsection{Roles and Objectives}
\label{subsec:roles}
We model the self-play training pipeline as an asymmetric two-role system. The Proposer generates problems. The Solver answers them. Both roles operate over the same task space $\mathcal{T}$, but they optimize different objectives.

Let $\pi_\phi$ denote the Proposer policy and $\pi_\theta$ the Solver policy \footnote{Following common practice in AZR-style frameworks, our implementation uses a single model with two prompt templates rather than two independent networks.}. At each outer step $t$:

\begin{enumerate}
    \item The Proposer samples a task $\tau = (q, \cdot)$ from $\pi_\phi(\cdot \mid c^{\text{prop}})$, where $c^{\text{prop}}$ is the proposer prompt template and $q = (\texttt{program}, \texttt{input})$ is a candidate problem.
    \item A deterministic environment $E$ executes the program on the input. If execution succeeds, $E$ returns the ground-truth output $o^*(q)$. The full task tuple is $\tau = (q, o^*(q))$.
    \item A data gate $F : \mathcal{T} \to \{0, 1\}$ decides whether $\tau$ enters the training pool $\mathcal{D}_t$. We formalize $F$ in Section~\ref{subsec:gate}.
    \item The Solver samples a candidate answer $a \sim \pi_\theta(\cdot \mid c^{\text{solv}}(q))$, where $c^{\text{solv}}$ is the solver prompt template containing $q$.
    \item Both roles receive rewards (in Section~\ref{subsec:reward}) and the policy parameters are updated jointly via GRPO.
\end{enumerate}

\paragraph{Optimization objectives.} The Solver maximizes expected reward over tasks drawn from the training pool:
\begin{equation}
\mathcal{J}_S(\theta) = \mathbb{E}_{\tau \sim \mathcal{D}_t, a \sim \pi_\theta(\cdot \mid c^{\text{solv}}(q))} \big[ R_S(a, \tau) \big].
\label{eq:solver-obj}
\end{equation}
The Proposer maximizes the difficulty its tasks pose to the current Solver:
\begin{equation}
\mathcal{J}_P(\phi) = \mathbb{E}_{\tau \sim \pi_\phi(\cdot \mid c^{\text{prop}})} \big[ R_P(\tau, \pi_\theta) \big].
\label{eq:proposer-obj}
\end{equation}

The asymmetry is structural. The Proposer optimizes a curriculum-shaping objective. The Solver optimizes a task-completion objective. These two objectives are coupled only through the training pool $\mathcal{D}_t$ and the dependence of $R_P$ on $\pi_\theta$.

\subsection{Reward Signals}
\label{subsec:reward}
\paragraph{Solver reward.}
Let $A(q) = a^{(1)}, \ldots, a^{(n)}$ denote the GRPO rollout group of size $n$ for a single solver prompt $c^{\text{solv}}(q)$. We use $n = 16$ in all experiments.

The \textit{grounded} solver reward checks each answer against the executor's ground-truth output:

\begin{equation}
R_S^{\text{g}}(a, \tau) = \mathbb{1}\big[  \texttt{eval}(a) = \texttt{eval}(o^*(q))  \big].
\label{eq:grounded-reward}
\end{equation}

The \textit{intrinsic} reward replaces the executor reference with intra-group agreement. Let $\kappa(a)$ be a canonical form of $a$. For each answer $a^{(i)}$ in the group:
\begin{equation}
R_S^{\text{i}}(a^{(i)}, A(q)) = \frac{1}{n} \sum_{j=1}^{n} \mathbb{1}\big[ \kappa(a^{(j)}) = \kappa(a^{(i)}) \big].
\label{eq:intrinsic-reward}
\end{equation}
The reward equals the answer's agreement share within the group. It is continuous in $[0, 1]$. A response that agrees with $k$ out of $n$ rollouts (including itself) earns reward $k/n$. The minimum is $1/n$ for a singleton; the maximum is $1$ when all rollouts agree.

\paragraph{Proposer reward.}
The Proposer reward depends on the Solver's pass rate on the Proposer's own task. Let $\hat{\alpha}(\tau, \pi_\theta)$ be the empirical Solver accuracy on $\tau$, estimated from $n_S = 8$ fresh Solver rollouts:
$$
\hat{\alpha}(\tau, \pi_\theta) = \frac{1}{n_S} \sum_{j=1}^{n_S} R_S^{\text{g}}\big(a^{(j)}, \tau\big), \qquad a^{(j)} \sim \pi_\theta(\cdot \mid c^{\text{solv}}(q)).
$$

The proposer reward is
$$
R_P(\tau) = 1 - \hat{\alpha}(\tau, \pi_\theta),
$$
The \textit{grounded} and \textit{intrinsic} Proposer variants differ only in what serves as the gold output during this estimation:
\begin{itemize}
    \item \textit{Grounded} Proposer: $\hat{\alpha}$ uses the executor output $o^*(q)$ as the reference.
    \item \textit{Intrinsic} Proposer: $\hat{\alpha}$ uses the Proposer's own self-claimed output $\tilde{o}(q)$ embedded in the prompt response, instead of $o^*(q)$.
\end{itemize}

\subsection{Data Gating}
\label{subsec:gate}
The system maintains an online training pool $\mathcal{D}_t$ of admitted tasks. Each step, the Proposer's candidates enter $\mathcal{D}_t$ only if they pass the data gate $F$. The Solver samples its training batch from a recent-biased mixture of the full pool, not from the latest batch alone.

The gate $F : \mathcal{T} \to \{0, 1\}$ determines whether a proposed task is admitted. Two natural endpoints define a binary design space: an \textbf{exec} mode admits $\tau$ only if the program executes successfully and produces a deterministic output across two repeated runs; an \textbf{off} mode admits every parseable response. We replace this binary choice with a one-parameter family. Let $\text{exec}(\tau) = 1$ when the executor produces a deterministic output and $0$ otherwise. Define
\begin{equation}
F_\varepsilon(\tau) = \begin{cases} 1 & \text{if } \text{exec}(\tau) = 1, \\ \xi_\tau & \text{if } \text{exec}(\tau) = 0, \end{cases} \qquad \xi_\tau \sim \text{Bernoulli}(\varepsilon).
\label{eq:gate}
\end{equation}
$\varepsilon$ controls the fraction of execution-failed tasks that the gate admits. We sweep $\varepsilon$ in Section~\ref{sec:phase-transitions} to characterize the phase transition between stable and collapsing regimes.

\subsection{Tasks}
\label{subsec:tasks}

We instantiate the formal pipeline of Sections~\ref{subsec:roles}--\ref{subsec:gate} on two tasks. The coding task asks the Solver to predict the printed output for a Proposer-supplied Python function and input. The DSL task has the same structure on a 15-operator prefix-expression language with two integer variables, where the output is always a single integer and the interpreter is deterministic. Appendix~B gives the full grammar.

The coding task is the natural setting for self-play on code and matches the design of prior work \citep{zhao2025absolute}. The DSL task is a controlled twin designed to remove three confounders present in the coding task: pretraining priors over the syntax, output-string ambiguity (floats, set order, side effects), and executor noise. Section~\ref{sec:policy-collapse} uses the DSL task to confirm that the observed collapse is driven by optimization dynamics, not by these environment-specific confounders.

\section{Training Collapse: Empirical Mechanisms}
\label{sec:policy-collapse}

This section presents the empirical evidence for the two-axis asymmetry introduced in Section~\ref{sec:formalization}. We make two claims. First, the data gate is the binding constraint on stability: removing it causes irreversible collapse regardless of reward design. Second, a proposer with access to ground-truth verification accelerates this collapse when paired with an intrinsic solver, a dynamic we call the \emph{Grounded Proposer Paradox}. We present results on coding in the main text. A controlled DSL twin task replicates the key findings and rules out confounders present in the coding task; the full DSL analysis is in Appendix~C.

\paragraph{Experimental matrix.} We run 7 configurations on each task, varying proposer reward, solver reward, and gate setting. Table~\ref{tab:matrix} lists the full matrix with results on both tasks. Run labels follow the convention \textbf{PS+gate}, where \textbf{P} is the proposer reward (G~=~grounded, I~=~intrinsic), \textbf{S} is the solver reward, and \textbf{gate} is \texttt{exec} or \texttt{off}. For coding we report in-domain validation accuracy (pretrained baseline $\approx 0.14$). For DSL we report a stratified offline holdout (pretrained baseline $\approx 0.53$; details in Appendix~C).

\begin{table}[t]
\centering
\small
\begin{tabular}{c|c|c|c|c|c|c}
\toprule
Label & Proposer & Solver & Gate & Coding & DSL & Outcome \\
\midrule
GG+exec & grounded & grounded & exec & $0.71$ & $0.61$ & stable \\
GI+exec & grounded & intrinsic & exec & $0.67$ & $0.63$ & stable \\
II+exec & intrinsic & intrinsic & exec & $0.67$ & $0.60$ & stable \\
GG+off & grounded & grounded & off & $0.002$ & $0.50$ & coding: collapse; DSL: baseline \\
IG+off & intrinsic & grounded & off & $0.006$ & $0.58$ & coding: collapse; DSL: baseline \\
GI+off & grounded & intrinsic & off & $0.002$ & $0.38$ & collapse \\
II+off & intrinsic & intrinsic & off & $0.007$ & $0.18$ & collapse \\
\bottomrule
\end{tabular}
\caption{Experimental matrix on coding and DSL tasks. Gate-on runs learn on both tasks. Gate-off runs collapse on coding regardless of reward design. On DSL, only intrinsic-solver cells collapse; grounded-solver cells hold at baseline because the DSL interpreter cannot produce ambiguous outputs.}
\label{tab:matrix}
\end{table}

\begin{figure}[t]
\centering
\includegraphics[width=\linewidth]{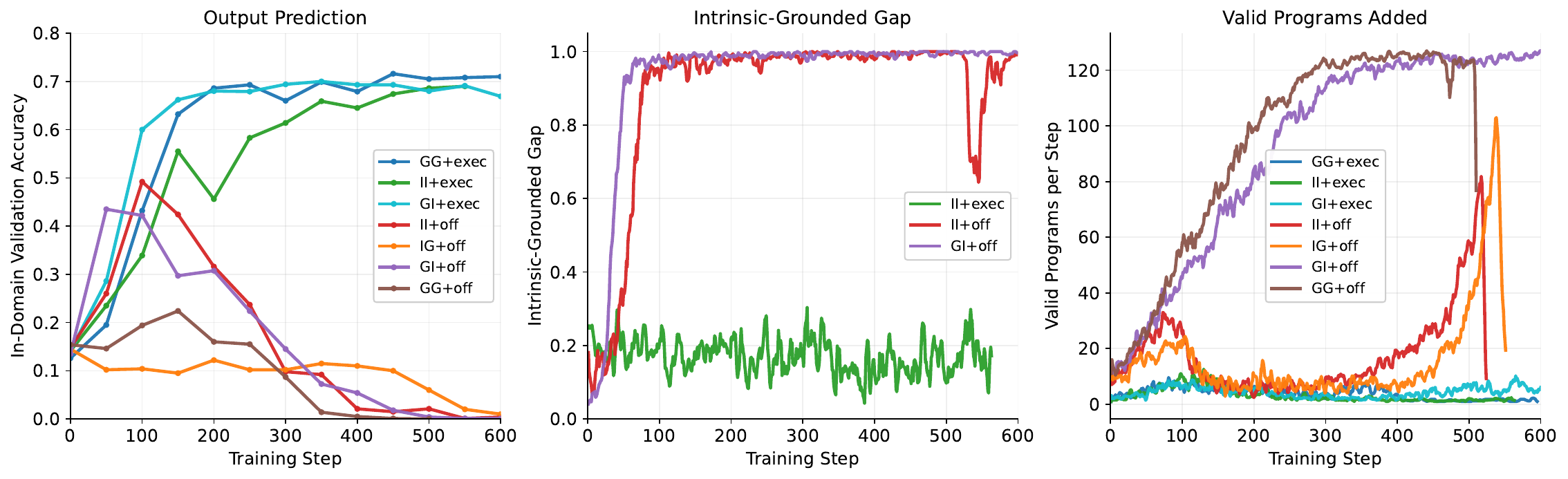}
\caption{Experimental overview on the coding task. \emph{Left:} in-domain validation accuracy. Gate-on runs learn; gate-off runs collapse to near zero. \emph{Center:} intrinsic-grounded gap for intrinsic-solver runs. II+off and GI+off saturate at gap $\approx 1.0$; II+exec stays near zero. \emph{Right:} per-step valid programs admitted to the training pool. Pretrained baseline: $0.14$.}
\label{fig:overview-code}
\end{figure}

\subsection{The Data Gate as the Binding Constraint}
\label{subsec:gate-as-constraint}

Every gate-off coding configuration falls to near zero regardless of reward design (Table~\ref{tab:matrix}). Every gate-on configuration learns. The gate is the binding constraint on stability.

The intrinsic-solver cells (II+off, GI+off) collapse through a self-consistency attractor. Without the gate, GRPO's group-relative advantages make intra-group consensus the cheapest direction in policy space, regardless of correctness. Once all rollouts agree on any canonical answer, the self-consistency reward is maximized and the gradient vanishes. The optimizer has reached an attractor that is consistent with the reward but inconsistent with the task. This is reward hacking in the classical sense \citep{karwowski2024goodhart}, but with a closed-loop twist: the spurious attractor propagates through the training pool to the proposer.

We measure this decoupling through the \emph{intrinsic-grounded gap}: the difference between the solver's self-consistency reward and its grounded accuracy. For II+off and GI+off, the gap saturates at $\approx 1.0$ within roughly 200 steps (Figure~\ref{fig:overview-code}, center panel). Restoring the gate (\textbf{II+exec}) keeps the gap near zero. The gate alone is sufficient to prevent the attractor.

The grounded-solver cells (GG+off, IG+off) collapse through a different route. Without the gate, the proposer emits programs whose output depends on hash randomization, floating-point display, or side effects. The training pool fills with these ambiguous programs and the grounded reward becomes uninformative. Even a grounded solver cannot learn from data that the gate would have rejected.

\begin{figure}[t]
\centering
\includegraphics[width=\linewidth]{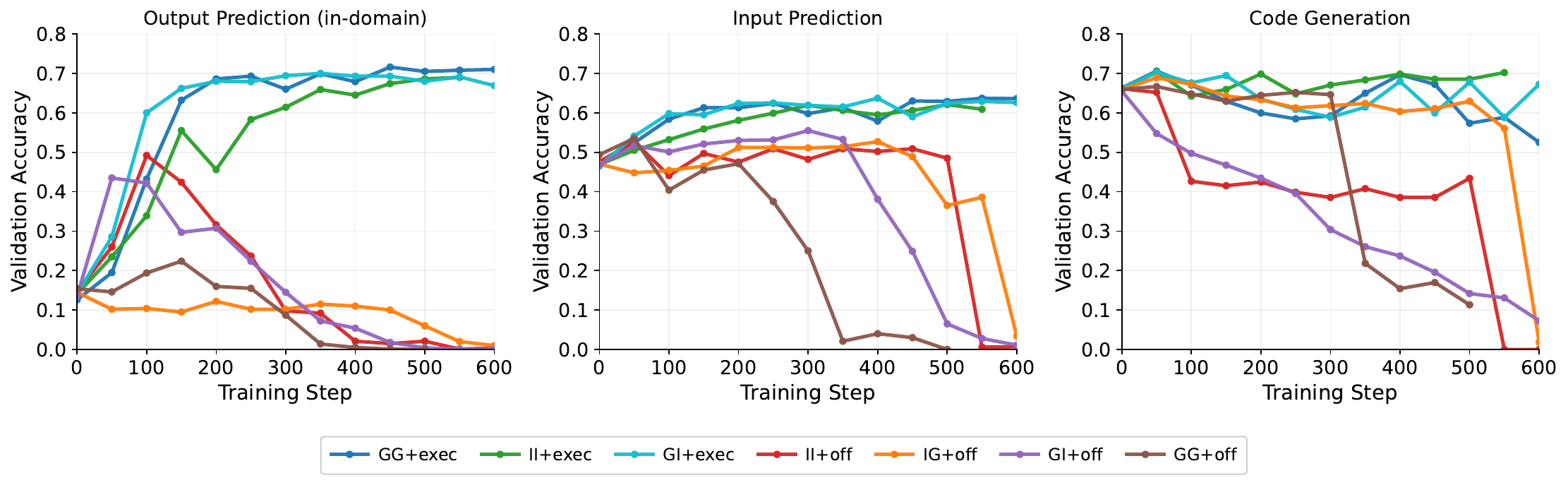}
\caption{Validation accuracy on three benchmarks: output prediction, input prediction, and code generation. Gate-on runs improve on all three; gate-off runs collapse uniformly. The gate's effect is not specific to the training objective.}
\label{fig:code-val-3benchs}
\end{figure}

The collapse generalizes beyond the in-domain metric. Figure~\ref{fig:code-val-3benchs} tracks all runs on three evaluation tasks: output prediction (in-domain), input prediction, and code generation. The three evaluation datasets are derived from CRUXEval-O, CRUXEval-I \citep{gu2024cruxeval}, and HumanEval+/MBPP+\citep{evalplus}, respectively. Gate-on runs improve on all three. Gate-off runs collapse on all three.

\paragraph{Identification.} The DSL twin task replicates the intrinsic-solver collapse under identical conditions, ruling out pretraining priors, output ambiguity, and executor noise as causes (Appendix~C). On DSL, where the interpreter is deterministic, the grounded-solver cells do not collapse. The contrast is informative: without the gate, intrinsic-solver collapse occurs on both tasks, so it is driven by optimization dynamics; grounded-solver collapse occurs only on coding, so it is driven by the environment producing ambiguous programs. The gate prevents both failure modes.

This contrast clarifies what the gate provides. On DSL, the deterministic interpreter already guarantees unambiguous outputs by construction, so the explicit gate is redundant for grounded solvers: the environment itself acts as an implicit gate. On real environments such as Python execution, no such guarantee exists, and all four gate-off cells collapse regardless of reward design. The function the gate provides (ensuring that only deterministic, unambiguous tasks enter the training pool) is necessary in both settings. The explicit filter is only redundant when the environment already provides this guarantee by construction.

\subsection{The Grounded Proposer Paradox}
\label{subsec:grounded-proposer-paradox}

\textbf{GI+off} pairs an intrinsic solver with a grounded proposer. \textbf{II+off} is the same but with an intrinsic proposer. The two runs differ only in proposer reward. GI+off collapses \emph{faster}: it reaches the self-consistent attractor in fewer steps (Figure~\ref{fig:overview-code}, left and center panels). Both runs end near zero on the in-domain metric (GI+off $0.002$, II+off $0.007$). The speed difference is in the trajectory, not the endpoint. The corresponding gate-on counterfactual, \textbf{GI+exec}, keeps validation accuracy high under the same grounded proposer and intrinsic solver, showing that the paradox is activated by removing the data gate rather than by grounded proposal itself.

\paragraph{Mechanism.} A grounded proposer concentrates its output on syntactically clean, semantically sharp programs. For an intrinsic solver, these tasks form the lowest-resistance path to the self-consistent attractor of Section~\ref{subsec:gate-as-constraint}. The proposer does not bias the solver toward truth. It sharpens the corridor through which the solver reaches the spurious fixed point. This is distinct from classical Goodhart failure \citep{karwowski2024goodhart, kwa2024catastrophic}, where a fixed proxy is exploited by a single optimizer. Here both agents co-evolve, and the upstream agent actively manufactures the structured inputs that accelerate the downstream failure.

\paragraph{Scope.} The paradox requires an intrinsic solver. With a grounded solver, GG+off still collapses on coding, but through the ambiguous-program route of Section~\ref{subsec:gate-as-constraint}, not through the self-consistency attractor. The proposer accelerates an already-doomed trajectory; it does not cause it.

Section~\ref{sec:phase-transitions} replaces the binary gate with the continuous leak rate $\varepsilon$ and traces the boundary between stable and collapsing regimes.

\section{Phase Transitions under Continuous Gate Relaxation}
\label{sec:phase-transitions}

Section~\ref{sec:policy-collapse} established that the execution gate is the binding constraint on stability in self-play. This section replaces the binary gate (on/off) with the continuous leak rate $\varepsilon$ introduced in Section~\ref{subsec:gate} and traces how system behavior changes as $\varepsilon$ increases from $0$ to $1$. Two results emerge. First, a sharp phase transition separates a stable regime from a collapsing regime, and the transition has a two-stage structure in which training-side decoupling appears at much lower $\varepsilon$ than validation-side collapse. Second, even in the stable regime, the system reaches a learning ceiling imposed by the proposer's finite capacity to generate novel problems.

\subsection{Phase Diagram of Leakage Rate $\varepsilon$}

\begin{figure}[t]
\centering
\includegraphics[width=\linewidth]{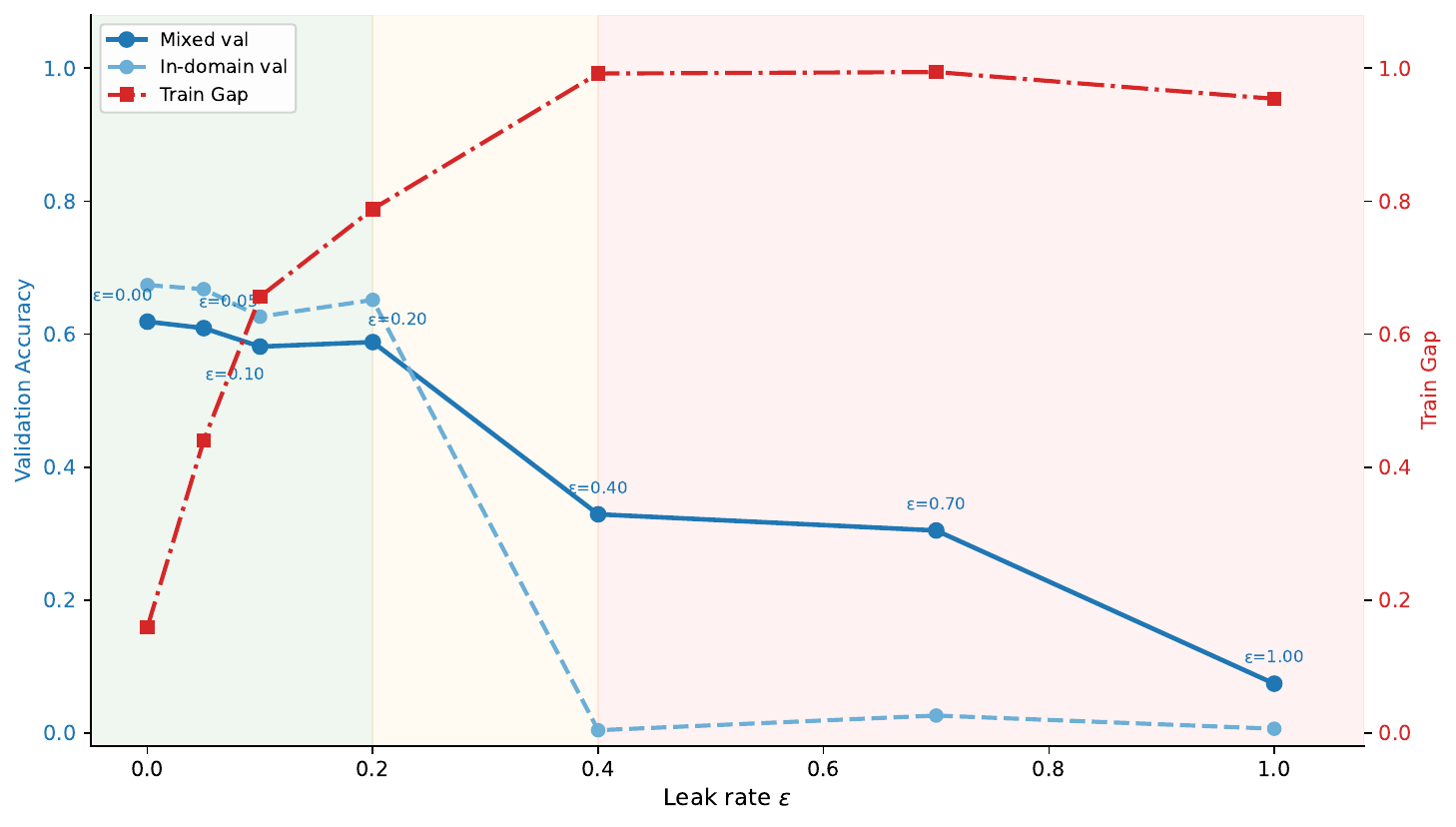}
\caption{Phase transition under increasing gate leak rate $\varepsilon$ (II configuration). Blue solid: mixed validation aggregate. Blue dashed: in-domain validation accuracy. Red dash-dot: training-side reward-grounded gap. Training-side decoupling begins at low $\varepsilon$. The in-domain probe reveals earlier hidden collapse at $\varepsilon = 0.40$, while the mixed aggregate remains near baseline until heavier corruption.}
\label{fig:phase-transition}
\end{figure}

We run the II configuration (intrinsic solver, intrinsic proposer) at seven values of $\varepsilon \in \{0.00, 0.05, 0.10, 0.20, 0.40, 0.70, 1.00\}$ and report late-stage means for two validation views (a mixed aggregate and an in-domain probe), the train-side gap, and train grounded accuracy. Figure~\ref{fig:phase-transition} plots all three against $\varepsilon$.

\paragraph{Two-stage structure.} The transition from stable to collapsed is not smooth. Two thresholds are visible, but the validation threshold depends on the evaluation view:

\begin{enumerate}
    \item \textbf{Training-side decoupling ($\varepsilon^*_{\text{train}} \approx 0.05$).} At $\varepsilon = 0.05$ the train gap jumps from $0.16$ to $0.44$. The solver's intra-group consensus begins to diverge from grounded accuracy on the training data. Yet validation accuracy barely moves: $0.674 \to 0.668$.
    \item \textbf{Validation-side degradation depends on the metric.} The in-domain probe collapses early: between $\varepsilon = 0.20$ and $\varepsilon = 0.40$, in-domain accuracy drops from $0.651$ to $0.004$. The mixed aggregate degrades later: it remains near baseline at $\varepsilon = 0.40$ ($0.329$ vs.\ step-0 $0.313$) and falls below baseline only by $\varepsilon = 0.70$ ($0.305$ vs.\ step-0 $0.316$). The aggregate metric masks a hidden failure mode that the in-domain probe already detects at moderate corruption.
\end{enumerate}

The gap between these thresholds is the central observation of this section. Moderate gate corruption ($\varepsilon \leq 0.20$) causes training-level decoupling while both validation views remain above their step-0 baselines. By $\varepsilon = 0.40$, the in-domain metric has collapsed but the mixed aggregate still appears robust. Only heavier corruption ($\varepsilon \geq 0.70$) pushes the aggregate view into decline. The mixed aggregate averages over easier and off-domain cases, so it can remain above baseline even after the in-domain probe has begun to deteriorate.

\paragraph{Connection to Youden's index.} The gate can be viewed as a binary classifier that retains valid programs (true positives) and rejects invalid ones (true negatives). Under our perturbation, the false positive rate equals $\varepsilon$: invalid programs are admitted with probability $\varepsilon$, while valid programs are always admitted. Youden's index $J = \text{TPR} - \text{FPR} = 1 - \varepsilon$. \citet{rad2026rate} prove that RLVR under a noisy oracle transitions from bounded learning to anti-learning when $J$ falls below zero. In our setting $J > 0$ for all $\varepsilon < 1$, so the Youden threshold alone does not predict either the earlier in-domain collapse at $\varepsilon = 0.40$ or the later aggregate degradation by $\varepsilon = 0.70$. \textbf{\textit{The self-play loop tightens the effective threshold}}: contaminated tasks enter the training pool, shift the proposer's distribution, and create a feedback cycle that amplifies noise beyond what a static-task analysis would predict. The effective critical $J$ in self-play is substantially higher than zero.

\subsection{The Proposer-Capacity Ceiling}

\begin{wrapfigure}{r}{0.48\linewidth}
\vspace{-2ex}
\centering
\includegraphics[width=\linewidth]{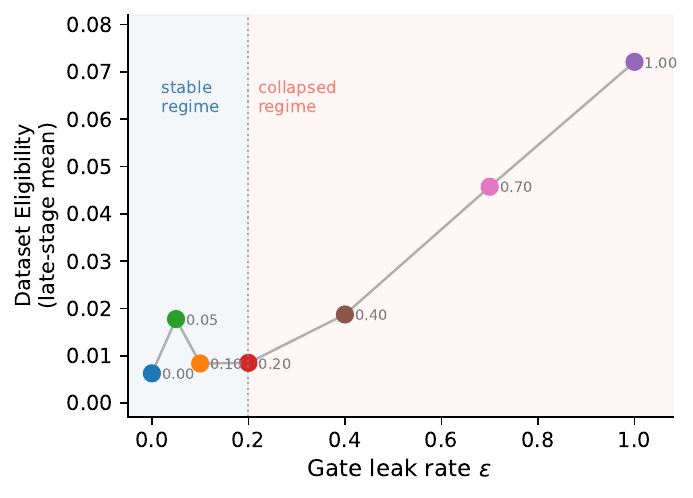}
\caption{Dataset eligibility vs.\ gate leak rate $\varepsilon$. Flat in the stable regime ($\varepsilon \leq 0.20$); rises only with training collapse, reflecting solver degradation rather than exploration.}
\label{fig:eligibility-vs-eps}
\vspace{-2ex}

\end{wrapfigure}

Even in the stable regime, the system reaches a learning ceiling. For $\varepsilon = 0.00$, late-stage dataset eligibility drops to $\approx 0.007$: fewer than $1\%$ of proposer outputs enter the training pool. The GG+exec trajectory (Figure~\ref{fig:overview-code}, left panel) shows this as a three-phase dynamic: rapid learning while the proposer contributes new problems (steps 1--160), slower gains from replay of pooled data (steps 160--440), and stagnation once replay is exhausted.

A natural conjecture is that a small non-zero $\varepsilon$ trades signal quality for a broader pool, delaying stagnation. The phase diagram does not support this. $\varepsilon = 0.00$ attains the best late-stage score on both validation views (mixed $0.619$, in-domain $0.674$). Dataset eligibility stays flat across the stable regime ($\varepsilon \leq 0.20$) and rises only where training has already collapsed (Figure~\ref{fig:eligibility-vs-eps}), confirming that the solver's difficulty-matching window, not the gate, is the binding constraint on pool size.

\begin{wrapfigure}{r}{0.48\linewidth}
\vspace{-2ex}
\centering
\includegraphics[width=\linewidth]{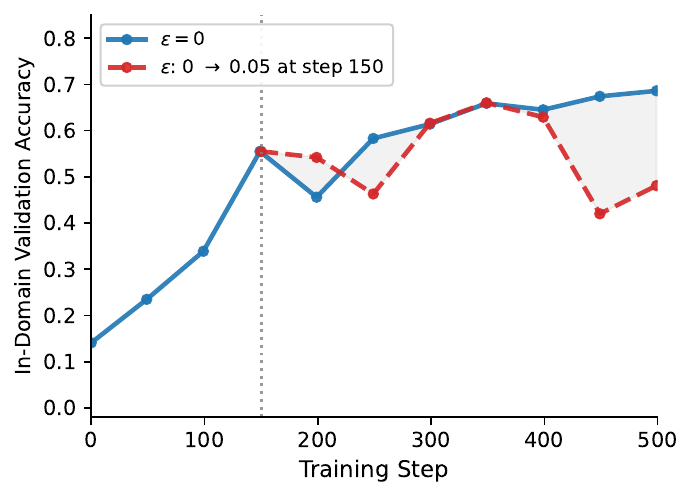}
\caption{Adaptive schedule: strict gate ($\varepsilon = 0$) for $150$ steps, then $\varepsilon = 0.05$ from the same checkpoint.}
\label{fig:adaptive-schedule}
\vspace{-2ex}
\end{wrapfigure}

\paragraph{Adaptive schedule.} A timing confound remains: each $\varepsilon$ in the phase diagram is applied from step $0$, so failure could be attributed to noisy gradients early in training. To rule this out, we trained with $\varepsilon = 0$ for $150$ steps and then resumed with $\varepsilon = 0.05$. Figure~\ref{fig:adaptive-schedule} shows the result. Through step $400$ the scheduled run tracks the baseline. By step $450$ its in-domain accuracy has dropped to $0.420$ while the baseline reaches $0.674$, and the gap persists at step $500$ ($0.481$ vs $0.686$). Gate relaxation does not help even after the model has stabilized.

\paragraph{Practical implication.} The strict gate ($\varepsilon = 0$) is optimal on both validation views we measured. The proposer-capacity ceiling is real but should be addressed through orthogonal means (curriculum design, periodic seeding with external tasks) rather than through the validity gate.

\section{Related Work}
\label{sec:related-work}
\subsection{Self-Play with Verifiable Rewards}
RLVR with rule-based environments and GRPO has driven gains in logical consistency \citep{deepseek2025r1}, but verifiable benchmarks are finite, motivating self-play approaches that grow their own data. Absolute Zero closes this loop with a co-evolving generator and solver \citep{zhao2025absolute}. Extensions target retrieval \citep{lu2025search}, policy diversity \citep{liang2025beyond}, long-context reasoning \citep{yang2025spell}, and self-generated reward \citep{simonds2025rlsr}.

The most closely related work is Self-Guided Self-Play \citep{bailey2026scaling}, developed concurrently. It identifies a Conjecturer reward-hacking failure and proposes a third Guide role that scores problems for relevance. The Guide is a policy-side soft signal during gradient computation. Our intervention is at a different layer: a hard filter on the data flow before the training pool. The two are orthogonal and could be combined.

\subsection{Reward Hacking and Goodhart's Law}
Reward hacking is a known failure mode in RL fine-tuning \citep{karwowski2024goodhart}, and KL regularization can fail under heavy-tailed misspecification \citep{kwa2024catastrophic}. In language models, corrupted reward signals activate memorization shortcuts \citep{yan2026spurious}, and code generation exhibits multi-phase rebound dynamics under reward hacking \citep{wu2026rebound}. ACE penalizes overconfident trajectories to stabilize optimization \citep{xu2026ace}.

\citet{he2026howfar} analyze unsupervised RLVR and argue that intrinsic methods succeed when the model's prior confidence aligns with correctness and fail when it does not. Their account locates the cause in the pretraining prior. Ours is complementary: we hold the prior fixed and show that the gate leakage rate governs whether collapse occurs. The prior shapes which attractor the policy collapses toward; the gate decides whether the system reaches it. In the self-play setting, the proposer's curriculum can co-evolve to amplify the solver's spurious signal, the dynamic we call the Grounded Proposer Paradox (Section~\ref{sec:policy-collapse}).

\subsection{Stability under Noisy Verifiers}
\citet{rad2026rate} study RLVR under noisy oracles and prove a phase transition governed by Youden's index $J$: below $J = 0$ the system enters anti-learning. \citet{plesner2026imperfect} report a related empirical finding: a verifier with up to 15\% label noise recovers most of the clean baseline. Both study fixed task distributions.

Our setting moves the noise inside a live self-play loop, where contaminated tasks shift the proposer's distribution and create a feedback cycle that amplifies noise beyond static-task predictions. In our experiments $J > 0$ for all $\varepsilon < 1$, yet collapse occurs at $\varepsilon = 0.40$ (in-domain) and $\varepsilon = 0.70$ (aggregate), showing that the effective critical $J$ in self-play is substantially higher than zero. We also report a two-stage structure (training-side decoupling before validation-side collapse) that has no analogue in the static-task setting.

\section{Conclusion}
\label{sec:conclusion}
The central finding of this work is that self-play stability is determined at the data level, not the reward level. A strict data filter is sufficient to prevent collapse regardless of reward design. No reward signal we tested is sufficient to prevent collapse without the filter.

This shifts the design question for self-play systems. The current literature focuses on building better reward signals: momentum anchors, confidence penalties, and hacking detectors. Our results suggest that the more productive axis is the data pipeline. What enters the training loop matters more than how the optimizer scores it once admitted.

Several questions remain open. Our phase diagram shows a gap between training-side decoupling and validation-side collapse. Why does validation remain robust when the training signal is already degraded? We attribute this to a regularization effect of the replay buffer, but a formal account is missing. Whether the optimal filter strictness can be adapted during training, rather than set as a fixed hyperparameter, is a natural next step. Finally, the Grounded Proposer Paradox we identified may extend beyond self-play to any multi-agent system where one agent's output feeds another's training data.

\clearpage
\bibliography{references}

\clearpage
\appendix
\section{Training Configuration}

All reported runs share the configuration below unless otherwise noted. The configuration is implemented through a single hydra config file plus environment overrides; see the codebase for the exact files.

\subsection{Model and optimization}

\begin{table}[h]
\centering
\small
\begin{tabular}{c|c}
\toprule
Parameter & Value \\
\midrule
Base model & Qwen3-4B-Base \\
Learning rate & $10^{-6}$ \\
Optimizer & AdamW, default Adam betas \\
KL loss coefficient $\beta_{\text{KL}}$ & $10^{-2}$ \\
KL loss type & \texttt{low\_var\_kl} \\
Advantage estimator & GRPO with std normalization \\
PPO clip ratio & $0.2$ \\
PPO epochs per outer step & $1$ \\
Total training steps & up to $700$ \\
\bottomrule
\end{tabular}
\end{table}

\subsection{Self-play loop}

\begin{table}[h]
\centering
\small
\begin{tabular}{c|c}
\toprule
Parameter & Value \\
\midrule
Per-step prompt batch ($b_P, b_S$) & $8$ \\
GRPO rollout group size ($n_P, n_S$) & $16$ \\
Inner solver rollouts for difficulty estimation ($n_S^{\text{diff}}$) & $8$ \\
Rollout sampling temperature & $1.0$ \\
Max prompt length & $3072$ tokens \\
Max response length & $2048$ tokens \\
Training pool cap $B$ & $16{,}384$ \\
Buffer eviction & FIFO \\
Sampling strategy & recent + uniform mix from full buffer \\
Solver dataloader window & per-step recent batch + uniform fill \\
Initial seed pool $\lvert\mathcal{D}_0\rvert$ & $32$ for the coding task, $24$ for the DSL task \\
\bottomrule
\end{tabular}
\end{table}

\subsection{Gate variants}

\begin{table}[h]
\centering
\small
\begin{tabular}{c|c}
\toprule
Parameter & Value \\
\midrule
\texttt{program\_validity.mode} & \texttt{execution} (headline runs) or \texttt{execution\_noisy} ($\varepsilon$-gradient runs) \\
Determinism check & two repeated executions per program \\
\texttt{program\_validity.seed} & $1$ (RNG seed for the Bernoulli leak draws) \\
Sandbox & per-program subprocess with $10$ s timeout \\
\bottomrule
\end{tabular}
\end{table}

\subsection{Hardware and runtime}

\begin{table}[h]
\centering
\small
\begin{tabular}{c|c}
\toprule
Parameter & Value \\
\midrule
Device & $8 \times$ A100 40 GB (single node) \\
Tensor model parallel size & $2$ \\
Sequence parallel size & $4$ \\
PPO micro batch size per GPU & $1$ \\
Rollout GPU memory utilization & $0.4$ \\
Actor parameter offload & enabled \\
Actor optimizer offload & enabled \\
\bottomrule
\end{tabular}
\end{table}

\section{DSL Specification}
\label{subsec:dsl}

The DSL is a small prefix-expression language used as a controlled twin to the coding task in Section~\ref{sec:policy-collapse}.

\subsection{Grammar}

In BNF:

\begin{verbatim}
expr     ::= literal | variable | unop "(" expr ")"
           | binop "(" expr "," expr ")"
           | "ITE" "(" cond "," expr "," expr ")"
cond     ::= cmp "(" expr "," expr ")"
literal  ::= signed integer
variable ::= "x" | "y"
unop     ::= "NEG" | "ABS"
binop    ::= "ADD" | "SUB" | "MUL" | "DIV" | "MOD" | "MAX" | "MIN"
cmp      ::= "GT"  | "LT"  | "EQ"  | "GEQ" | "LEQ"
\end{verbatim}

Fifteen operators in total: seven binary arithmetic, two unary arithmetic, one conditional, five comparators.

\subsection{Operator semantics}

\begin{table}[h]
\centering
\small
\begin{tabular}{c|c}
\toprule
Operator & Semantics \\
\midrule
\texttt{ADD(a, b)} & $a + b$ \\
\texttt{SUB(a, b)} & $a - b$ \\
\texttt{MUL(a, b)} & $a \cdot b$ \\
\texttt{DIV(a, b)} & $\lfloor a / b \rfloor$ ($a$ \texttt{//} $b$ in Python). Rejected when $b = 0$. \\
\texttt{MOD(a, b)} & $a \bmod b$. Rejected when $b = 0$. \\
\texttt{MAX(a, b)} & $\max(a, b)$ \\
\texttt{MIN(a, b)} & $\min(a, b)$ \\
\texttt{NEG(a)} & $-a$ \\
\texttt{ABS(a)} & $\lvert a \rvert$ \\
\texttt{ITE(c, t, e)} & evaluates to $t$ if $c$ is true, else $e$ \\
\texttt{GT(a, b)} & $a > b$ \\
\texttt{LT(a, b)} & $a < b$ \\
\texttt{EQ(a, b)} & $a = b$ \\
\texttt{GEQ(a, b)} & $a \geq b$ \\
\texttt{LEQ(a, b)} & $a \leq b$ \\
\bottomrule
\end{tabular}
\end{table}

\subsection{Interpreter properties}

The interpreter is a pure-Python evaluator $E_{\text{DSL}} : G_{\text{DSL}} \times \mathbb{Z}^2 \to \mathbb{Z} \cup \{\bot\}$ where $G_{\text{DSL}}$ denotes the set of well-formed expressions and $\bot$ is the rejection symbol returned for runtime failures (such as division by zero on the supplied input).

By construction:

\begin{enumerate}
    \item Every accepted expression and input pair has exactly one output.
    \item The output is always a single integer or $\bot$.
    \item There are no floats, collections, randomness sources, or side effects.
    \item The interpreter is deterministic: repeated execution on the same input yields the same output.
\end{enumerate}

\subsection{Validation}

A candidate expression is admitted to the training pool only if the interpreter returns a non-$\bot$ value on every point of a $5 \times 5$ probe grid over $(x, y) \in [-2, 2]^2$. This is the DSL analogue of the coding execution gate. During seed task construction, variable ranges are extended to $x, y \in [-10, 10]$ to produce a richer initial pool.

The Solver's grounded reward on the DSL task is exact integer equality after whitespace stripping. We do not use Python \texttt{eval} for comparison, since the DSL has no value ambiguity (\texttt{"1"} and \texttt{"1.0"} cannot both be valid outputs because the output type is constrained to integer).

\subsection{Examples}

\begin{verbatim}
ADD(x, y)                                           ; x + y
ABS(SUB(x, y))                                      ; |x - y|
ITE(GT(x, 0), x, NEG(x))                            ; abs(x) via conditional
ADD(MUL(x, x), MUL(y, y))                           ; x^2 + y^2
ITE(EQ(MOD(x, 2), 0), DIV(x, 2), ADD(MUL(x, 3), 1)) ; one Collatz step
\end{verbatim}

\subsection{Differences from the coding task}

The DSL task is structurally identical to the coding task in the formal pipeline of Sections~\ref{subsec:roles}--\ref{subsec:gate}. The differences are confined to the executor and the answer-comparison rule:

\begin{table}[h]
\centering
\small
\begin{tabular}{c|c|c}
\toprule
Property & Coding task & DSL task \\
\midrule
Executor & Python interpreter in subprocess sandbox & Pure-Python recursive evaluator \\
Output type & any printable value (string) & single integer \\
Comparison & \texttt{eval} on both sides (type-tolerant) & exact integer equality \\
Pretraining prior & substantial & none \\
Determinism check & two repeated runs & by construction \\
Timeout & $10$ s & not needed \\
\bottomrule
\end{tabular}
\end{table}

These differences are exactly what Section~\ref{subsec:tasks} calls ``three coding-specific confounders by construction.''

\section{DSL Replication of Section~\ref{sec:policy-collapse} Results}

This appendix presents the full DSL results referenced from Section~\ref{sec:policy-collapse}. The DSL task removes three coding-specific confounders by construction: pretraining priors, output ambiguity, and executor noise. If the collapse mechanisms of Section~\ref{sec:policy-collapse} reproduce on DSL, they cannot be attributed to any of these confounders.

\subsection{Overview}

\begin{figure}[h]
\centering
\includegraphics[width=\linewidth]{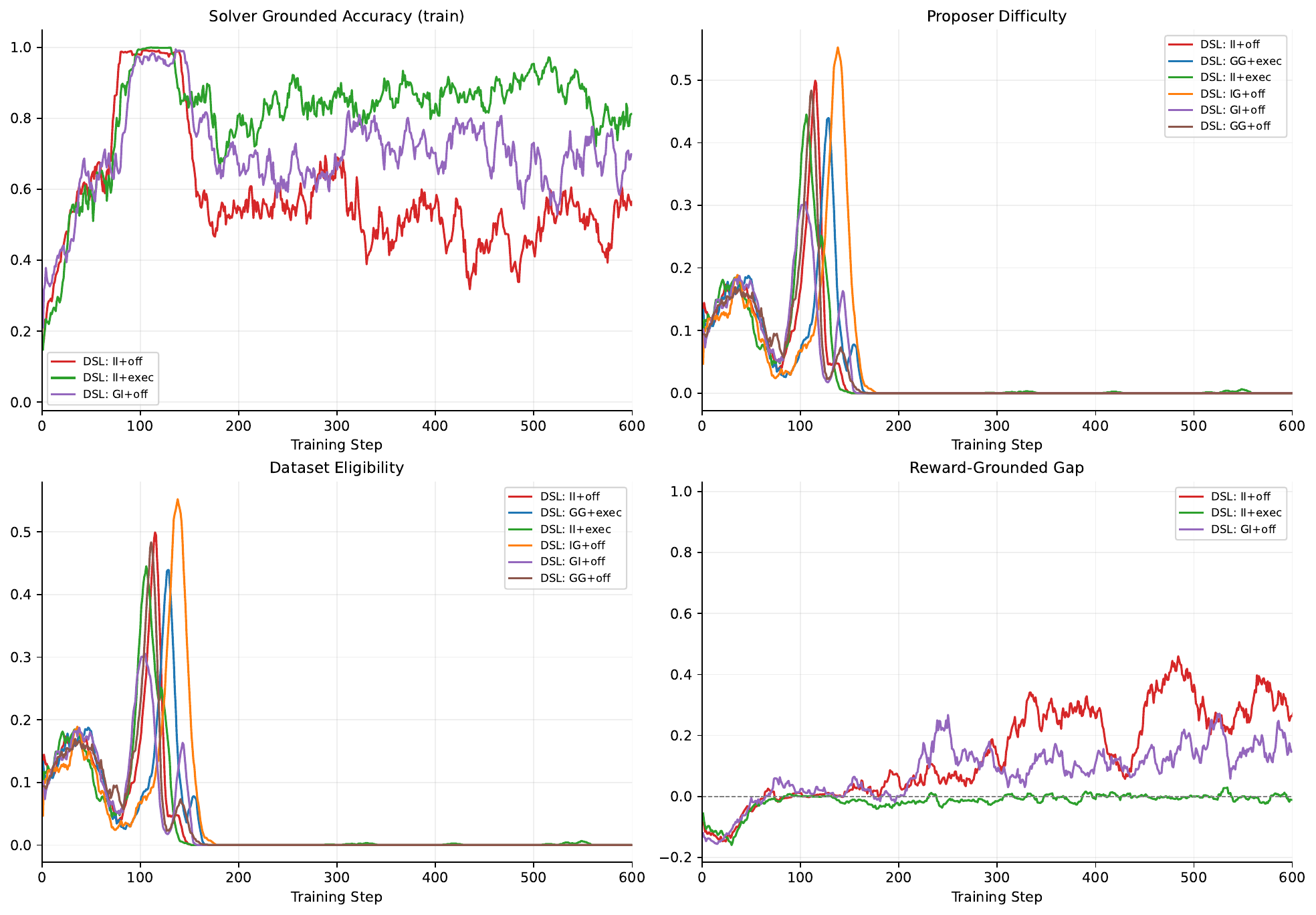}
\caption{DSL experimental overview. Top left: solver grounded accuracy (train). Top right: proposer difficulty. Bottom left: dataset eligibility. Bottom right: reward-grounded gap. Gate-on runs (GG+exec, II+exec) learn; intrinsic-solver off-gate runs (II+off, GI+off) show rising gap; grounded-solver off-gate runs (GG+off, IG+off) remain near baseline.}
\label{fig:dsl-overview}
\end{figure}

We run the same configurations as the coding task (Table~\ref{tab:matrix}). Figure~\ref{fig:dsl-overview} shows four diagnostic metrics across  DSL runs.

\subsection{Catastrophic decoupling replicates on DSL}

The bottom-right panel of Figure~\ref{fig:dsl-overview} shows the reward-grounded gap for the intrinsic-solver DSL runs. II+off shows a rising gap, confirming the same directional pattern as the coding task. The absolute magnitude is smaller because the DSL's deterministic integer output limits the spurious attractor's reach. II+exec stays near zero, confirming that the gate prevents decoupling on DSL as well.

\subsection{Grounded-solver cells do not collapse on DSL}

On the coding task, all four off-gate cells collapse (Section~\ref{subsec:gate-as-constraint}). On DSL, only the intrinsic-solver cells collapse. The grounded-solver cells (GG+off, IG+off) remain within sampling noise of the pretrained baseline ($\approx 0.53$ on the d4-6 tier). This is because the DSL interpreter is deterministic and cannot produce ambiguous programs. When the solver is grounded, the deterministic interpreter already provides the gate's function (unambiguous outputs), making the explicit filter redundant.

\subsection{Stratified holdout evaluation}

\begin{figure}[h]
\centering
\includegraphics[width=\linewidth]{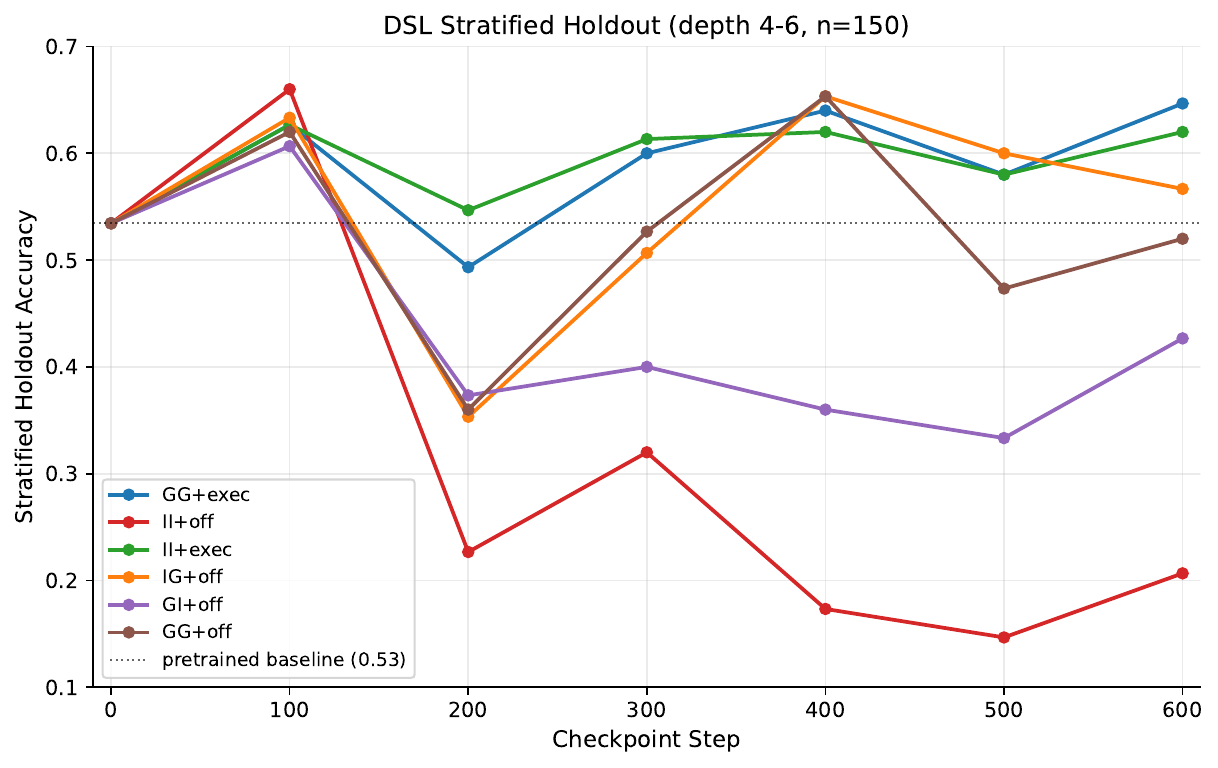}
\caption{DSL stratified holdout accuracy (depth 4-6) across training. Gate-on runs learn. Grounded-solver off-gate runs hold at baseline. Intrinsic-solver off-gate runs collapse below baseline.}
\label{fig:dsl-stratified}
\end{figure}

Figure~\ref{fig:dsl-stratified} reports the offline stratified holdout (depth 4-6, $n = 150$ per run) at checkpoint steps $0, 100, \ldots, 600$. The pretrained baseline is $\approx 0.53$. Gate-on runs (GG+exec, II+exec) rise to $\approx 0.60$. Grounded-solver off-gate runs stay near baseline. Intrinsic-solver off-gate runs fall below baseline, with II+off reaching $0.18$.

\end{document}